\documentclass[conference]{IEEEtran}
\IEEEoverridecommandlockouts
\usepackage{cite}
\usepackage{subcaption}
\usepackage{amsmath,amssymb,amsfonts}
\usepackage{algorithmic}
\usepackage{graphicx}
\usepackage{textcomp}
\usepackage{xcolor}
\usepackage{orcidlink}
\usepackage{hyperref}
\def\BibTeX{{\rm B\kern-.05em{\sc i\kern-.025em b}\kern-.08em
    T\kern-.1667em\lower.7ex\hbox{E}\kern-.125emX}}

\title{TGO-I: Spectral Geometry Observatory}
\author{
\IEEEauthorblockN{Kaustubh Kapil\orcidlink{0009-0000-4918-8452} and Kishor P. Upla\orcidlink{0000-0001-6306-0682}}
\IEEEauthorblockA{
Department of Electronics Engineering\\
Sardar Vallabhai National Institute of Technology (SVNIT), Surat, India\\
u24ec049@eced.svnit.ac.in, kishorupla@gmail.com
}
}%}

\begin{document}

\maketitle

\begin{abstract}
Despite the widespread adoption of ViT and its utilization in multiple computer vision applications, the fundamental understanding of their dimensional and representational geometry remains less explored as compared to their application adaptation. To address this gap, we introduce \textbf{Transformer Geometry Observatory (TGO)} which is a surgical set of experiments and analyses pipelines built to analyze the representational geometry and dynaimcs of ViT. \textbf{TGO-I} specifically looks into the \emph{Spectral Geometry} of ViT. It evaluates the ViT on the ImageNet-100 to analyze Effective Rank, Stable Rank, Participation Ratio, Spectral Entropy, Spectral Flatness, Spectral Anisotropy, covariance structure, Eigenspectra, and singular value spectra. Results show a consistent increase in dimensional utilization across training, accompanied by decreasing anisotropy, increasing spectral entropy, increasing participation ratio, and progressively flatter eigenspectra. Contrary to the common intuition that training should concentrate information into a small number of dominant directions, a progressive redistribution of variance across representational dimensions is observed. This effect is particularly pronounced in the final CLS token, which exhibits the highest effective dimensionality and lowest anisotropy in the network.
\end{abstract}

\section{Introduction}

Vision Transformers~\cite{ViT} (ViTs) have emerged as a dominant architecture in modern computer vision, achieving state-of-the-art performance across image classification, segmentation, detection, and multimodal learning tasks. While considerable effort has been devoted to understanding transformer behavior through attention analysis, feature visualization, pruning studies, representation similarity analysis, and downstream performance evaluation, a complete picture of how transformer representations evolve throughout training remains elusive. A growing body of work has explored representation geometry, intrinsic dimensionality, neural collapse, and spectral properties of deep neural networks. However, these analyses are often limited to specific layers, isolated training stages, or individual observables. As a result, our understanding of how representation geometry evolves jointly across network depth and training time remains incomplete. Understanding this evolution is important because representation geometry influences information capacity, redundancy, feature utilization, optimization dynamics, compression behavior, and interpretability. Questions such as how dimensional utilization changes during training, how variance is distributed across representational directions, and how global representations emerge within transformer architectures remain largely under-explored.

To investigate these questions, this study introduces Transformer Geometry Observatory (TGO), a research framework designed to systematically instrument and analyze the internal geometry of transformer representations. Rather than focusing on model performance or attention patterns alone, TGO treats transformer activations as evolving geometric objects and studies how their structure changes throughout training. This work presents the first installment of the framework, Transformer Geometry Observatory I (TGO-I), which focuses on spectral representation geometry. Using a ViT-Small/16 trained on ImageNet-100~\cite{imagenet}, we instrument patch embeddings, positional embeddings, transformer blocks, and the final CLS representation throughout training. Then their covariance structure, eigenspectra, singular value spectra, Effective Rank, Stable Rank, Participation Ratio, Spectral Entropy, Spectral Flatness, and Spectral Anisotropy are analyzed using a fixed analysis subset. Our experiments reveal a consistent increase in dimensional utilization throughout training, accompanied by increasing spectral entropy, increasing participation ratio, decreasing anisotropy, and progressively flatter eigenspectra. These observations indicate a progressive redistribution of variance across representational dimensions. The effect is particularly pronounced in the final CLS representation, which exhibits substantially higher effective dimensionality and lower anisotropy than intermediate transformer layers. The contributions of this work are summarized as follows:
\begin{enumerate}

    \item Transformer Geometry Observatory (TGO) is introduced as a modular framework for studying representation dynamics in transformer architectures.

    \item A spectral observatory is developed to track geometric evolution across layers and training epochs.

    \item Empirical evidence is presented showing that Vision Transformer representations become progressively less spectrally concentrated and increasingly distributed throughout training.

    \item The final CLS representation is identified as a uniquely distributed geometric object exhibiting the highest dimensional utilization and lowest anisotropy within the network.

    \item A foundation is established for future observatories aimed at studying token dynamics, representation similarity, attention behavior, redundancy, and bottleneck formation in transformer architectures.

%\end{enumerate}
\end{enumerate}

\subsection{Research Questions}

This work investigates the following questions:

RQ1: How does representation dimensionality evolve during ViT training?

RQ2: How is variance distributed across representational dimensions?

RQ3: How does representation geometry evolve across network depth?

RQ4: Does the CLS token exhibit distinct geometric behavior compared to intermediate representations?

RQ5: Are trends observed consistently across multiple spectral observables?

The complete implementation of TGO-I, trained checkpoints, observatory pipelines,
and generated artifacts are publicly available at:
\href{https://github.com/KaustubhKapil/Transformer_Geometry_Observatory_Part-1}{GitHub Repository}.

\section{Framework Methodology}

The primary objective of TGO-I is to characterize how the geometry of Vision Transformer representations evolves throughout training. Rather than analyzing model predictions or attention maps directly, TGO-I studies the statistical structure of internal representations through covariance-based spectral analysis. Covariance spectra provide a compact description of variance allocation across representational dimensions and enable the study of dimensional utilization, anisotropy, and spectral concentration. A ViT-Small/16 model was trained on the ImageNet-100 dataset for 100 epochs. Activations were extracted from the Patch Embedding layer, Positional Embedding layer, all Transformer blocks, and the final CLS representation using forward hooks. To ensure consistent observability throughout training, all measurements were performed on a fixed analysis subset consisting of 1000 validation images.

The feature covariance matrix for layer $l$ is computed as

\begin{equation}
\mathbf{C}_{l}
=
\frac{1}{N-1}
\left(
\mathbf{X}_{l}
-
\mathbf{1}\boldsymbol{\mu}_{l}^{T}
\right)^{T}
\left(
\mathbf{X}_{l}
-
\mathbf{1}\boldsymbol{\mu}_{l}^{T}
\right),
\end{equation}

where

\begin{equation}
N = B \times T
\end{equation}

and

\begin{equation}
\boldsymbol{\mu}_{l}
=
\frac{1}{N}
\sum_{i=1}^{N}
\mathbf{X}_{l}^{(i)}
\end{equation}

is the feature-wise mean representation.

Spectral analysis is then performed through eigendecomposition of the covariance matrix,

\begin{equation}
\mathbf{C}_{l}
=
\mathbf{V}_{l}
\mathbf{\Lambda}_{l}
\mathbf{V}_{l}^{T},
\end{equation}

where

\begin{equation}
\mathbf{\Lambda}_{l}
=
\mathrm{diag}
\left(
\lambda_{1},
\lambda_{2},
\ldots,
\lambda_{D}
\right)
\end{equation}
contains the covariance eigenvalues ordered such that

\begin{equation}
\lambda_{1}
\ge
\lambda_{2}
\ge
\cdots
\ge
\lambda_{D}.
\end{equation}
These eigenvalues form the foundation of all observables studied in TGO-I, including Effective Rank, Stable Rank, Participation Ratio, Spectral Entropy, Spectral Flatness, Spectral Anisotropy, Eigenspectra, Singular Value Spectra, and Spectral Decay. 

\begin{figure*}[h]
    \centering
    \includegraphics[width=0.9\linewidth]{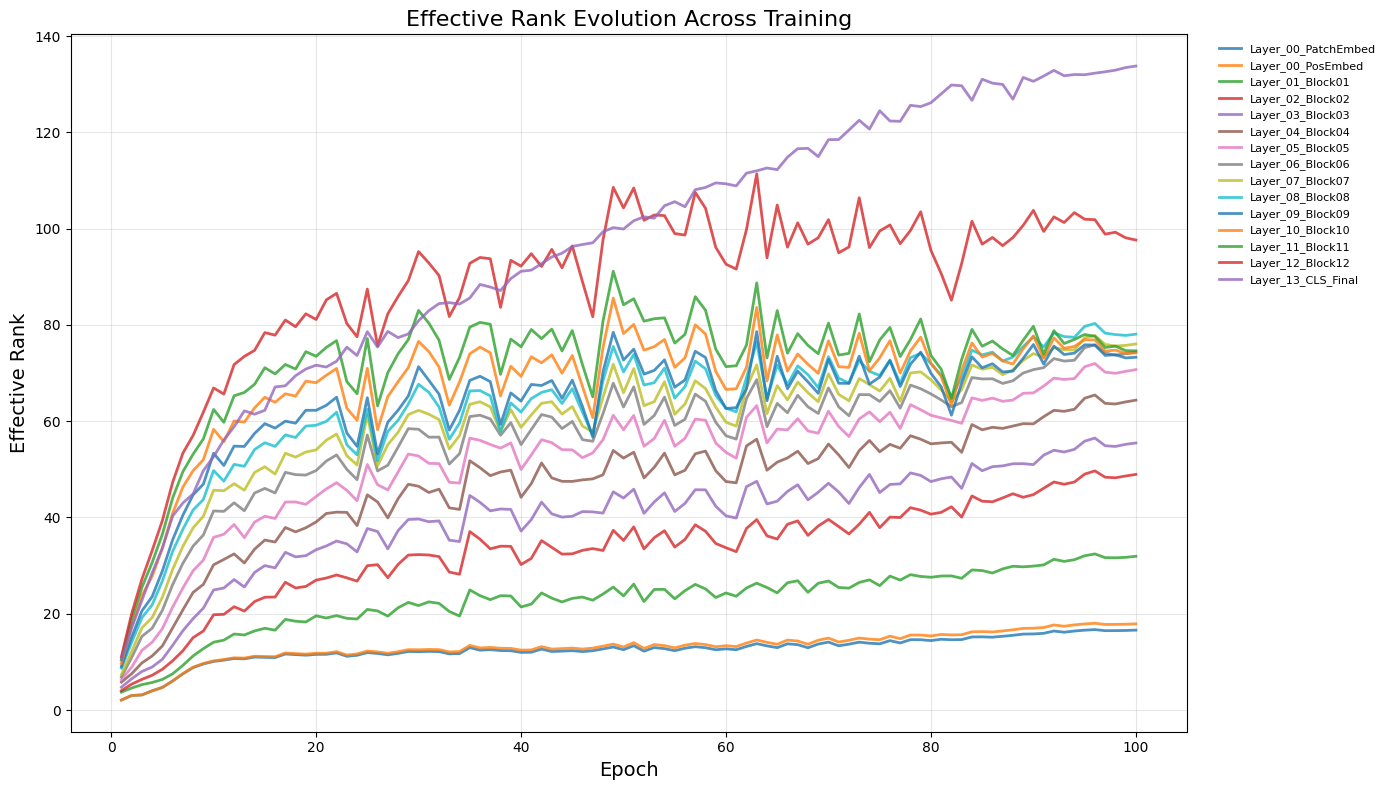}
    \caption{Effective Rank evolution across 100 epochs for all layers. It can be observed that over the course of training ViT, the rank increases. Positional and Patch embeddings show relative stability in rank. \textbf{CLS} shows constant rise in effective rank.}
    \label{fig:effective_rank_layers}
\end{figure*}

\section{Spectral Observables}

To characterize the evolution of representation geometry throughout training, TGO-I analyzes the eigenspectra of layer covariance matrices using a collection of complementary spectral observables. Each observable captures a distinct aspect of dimensional utilization, variance allocation, and spectral concentration.

Let the covariance matrix of layer $l$ be denoted by

\begin{equation}
\mathbf{C}_l \in \mathbb{R}^{D \times D}
\end{equation}

with eigenvalues

\begin{equation}
\lambda_1 \ge \lambda_2 \ge \cdots \ge \lambda_D \ge 0.
\end{equation}

The normalized eigenvalue distribution is defined as

\begin{equation}
p_i
=
\frac{\lambda_i}
{\sum_{j=1}^{D}\lambda_j}.
\end{equation}

\subsection{Effective Rank}

Effective Rank estimates the number of dimensions that meaningfully contribute to the representation. It is computed as the exponential of the entropy of the normalized eigenvalue distribution:

\begin{equation}
H
=
-\sum_{i=1}^{D}
p_i \log p_i
\end{equation}

\begin{equation}
r_{\mathrm{eff}}
=
\exp(H).
\end{equation}

Higher Effective Rank indicates that variance is distributed across a larger number of spectral directions.

\subsection{Stable Rank}

Stable Rank provides a robust estimate of dimensionality based on the ratio of total spectral energy to the dominant spectral direction:

\begin{equation}
r_{\mathrm{stable}}
=
\frac{\|\mathbf{C}\|_F^2}
{\|\mathbf{C}\|_2^2}
=
\frac{\sum_{i=1}^{D}\lambda_i^2}
{\lambda_1^2}.
\end{equation}

Higher Stable Rank indicates reduced dominance of the leading eigenvalue.

\subsection{Participation Ratio}

Participation Ratio estimates the effective number of dimensions contributing to the covariance spectrum:

\begin{equation}
PR
=
\frac{\left(\sum_{i=1}^{D}\lambda_i\right)^2}
{\sum_{i=1}^{D}\lambda_i^2}.
\end{equation}

Large values indicate that variance is spread across many dimensions, whereas small values indicate concentration within a limited subspace.

\subsection{Spectral Entropy}

Spectral Entropy quantifies the uniformity of the eigenvalue distribution:

\begin{equation}
H_{\mathrm{spectral}}
=
-\sum_{i=1}^{D}
p_i \log p_i.
\end{equation}

Higher values correspond to more evenly distributed variance across spectral directions.

\subsection{Spectral Anisotropy}

Spectral Anisotropy measures the dominance of the principal spectral direction relative to the total variance:

\begin{equation}
A
=
\frac{\lambda_1}
{\sum_{i=1}^{D}\lambda_i}.
\end{equation}

Large anisotropy values indicate strong concentration of variance within a small number of directions, whereas low anisotropy values indicate a more distributed spectral structure.

\subsection{Covariance Structure and Spectral Visualization}

In addition to scalar observables, TGO-I analyzes covariance matrices, eigenspectra, singular value spectra, and spectral decay curves directly. These visualizations provide qualitative insight into variance allocation, feature correlation structure, spectral decay behavior, and the emergence of dominant or distributed representational directions throughout training.

Together, these observables provide complementary measurements of representation geometry and enable the study of dimensional utilization, variance redistribution, and spectral evolution across both training time and network depth.

\section{Experiments}
This section describes the experimental configuration used to investigate the spectral evolution of Vision Transformer representations throughout training. A ViT-Small/16 model was trained on the ImageNet-100 dataset for 100 epochs using a NVIDIA Quadro RTX 6000 GPU. To ensure consistent observability, all spectral measurements were performed on a fixed analysis subset of 1000 validation images throughout training. The following subsections describe the model configuration, activation extraction pipeline, analysis subsets, and training parameters used throughout TGO-I.

\subsection{Model Configuration}

The model used throughout TGO-I was ViT-Small/16. The architecture consists of a patch embedding layer followed by twelve Transformer encoder blocks and a final classification head operating on the CLS token representation.

Input images were partitioned into non-overlapping $16 \times 16$ patches and projected into a 384-dimensional embedding space. A learnable CLS token and positional embeddings were added prior to Transformer processing. The resulting token sequence was propagated through twelve encoder blocks composed of Multi-Head Self-Attention and Feed-Forward Network modules.

For an input resolution of $224 \times 224$, the model produces a sequence of 197 tokens, comprising 196 patch tokens and one CLS token.

\subsection{Activation Extraction}

To observe representation evolution throughout training, forward hooks were registered at multiple stages of the network. Activations were extracted from:

\begin{itemize}
\item Patch Embedding Output
\item Positional Embedding Output
\item Transformer Block 1--12 Outputs
\item Final CLS Representation
\end{itemize}

For each layer, activations from the fixed analysis subset were aggregated to construct a representation matrix

\begin{equation}
\mathbf{X}_l \in \mathbb{R}^{(B \times T)\times D}
\end{equation}

where $B$ denotes the number of images, $T$ the number of tokens per image, and $D$ the embedding dimension.

These representations formed the basis for all covariance and spectral analyses performed in TGO-I.

\subsection{Analysis Set Construction}

A key design principle of TGO-I is observational consistency. Spectral observables were therefore never computed on training mini-batches. Instead, all measurements were performed on the fixed analysis subset described previously.

For ViT-Small/16,

\begin{equation}
B = 1000,\qquad
T = 197,\qquad
D = 384
\end{equation}

resulting in an aggregated representation matrix of size

\begin{equation}
\mathbf{X}_l
\in
\mathbb{R}^{197000 \times 384}.
\end{equation}

Tokens from all images were treated as observations, while embedding dimensions were treated as variables. This construction enables the study of dataset-level representation geometry rather than image-specific token dynamics.

\subsection{Training Configuration}

The ViT-Small/16 model was trained for 100 epochs using PyTorch with Automatic Mixed Precision (AMP) enabled. Model checkpoints were saved throughout training to enable longitudinal geometric analysis across epochs.

At the end of each epoch, the fixed analysis subset was processed through the network and activations were collected from all monitored layers. Covariance matrices were then constructed and analyzed using the spectral observables described in Section III.

This procedure produced a complete temporal record of representation geometry evolution across both network depth and training time, forming the basis of all results reported in TGO-I.

\section{Findings}

This section presents the primary observations obtained from the spectral observatory. The purpose of this section is to document measured behavior rather than propose explanations for the observed trends. Key findings include consistent plots for reference

\subsection{Effective Rank Evolution}
\textbf{Effective Rank} was analyzed per epoch for all layers. Figure~\ref{fig:effective_rank_layers} shows the evolution of Effective Rank throughout training. A consistent increase in Effective Rank was observed across nearly all monitored layers. Early representations, including the Patch Embedding and Positional Embedding layers, remained comparatively low-dimensional throughout training, whereas deeper Transformer layers exhibited progressively larger Effective Rank values. The strongest increase was observed in the final CLS representation. Across the network, Effective Rank increased substantially between initialization and epoch 100, indicating progressively greater dimensional utilization of the available feature space.

\begin{figure*}
    \centering

    \begin{subfigure}[b]{0.9\textwidth}
        \centering
        \includegraphics[width=\linewidth]{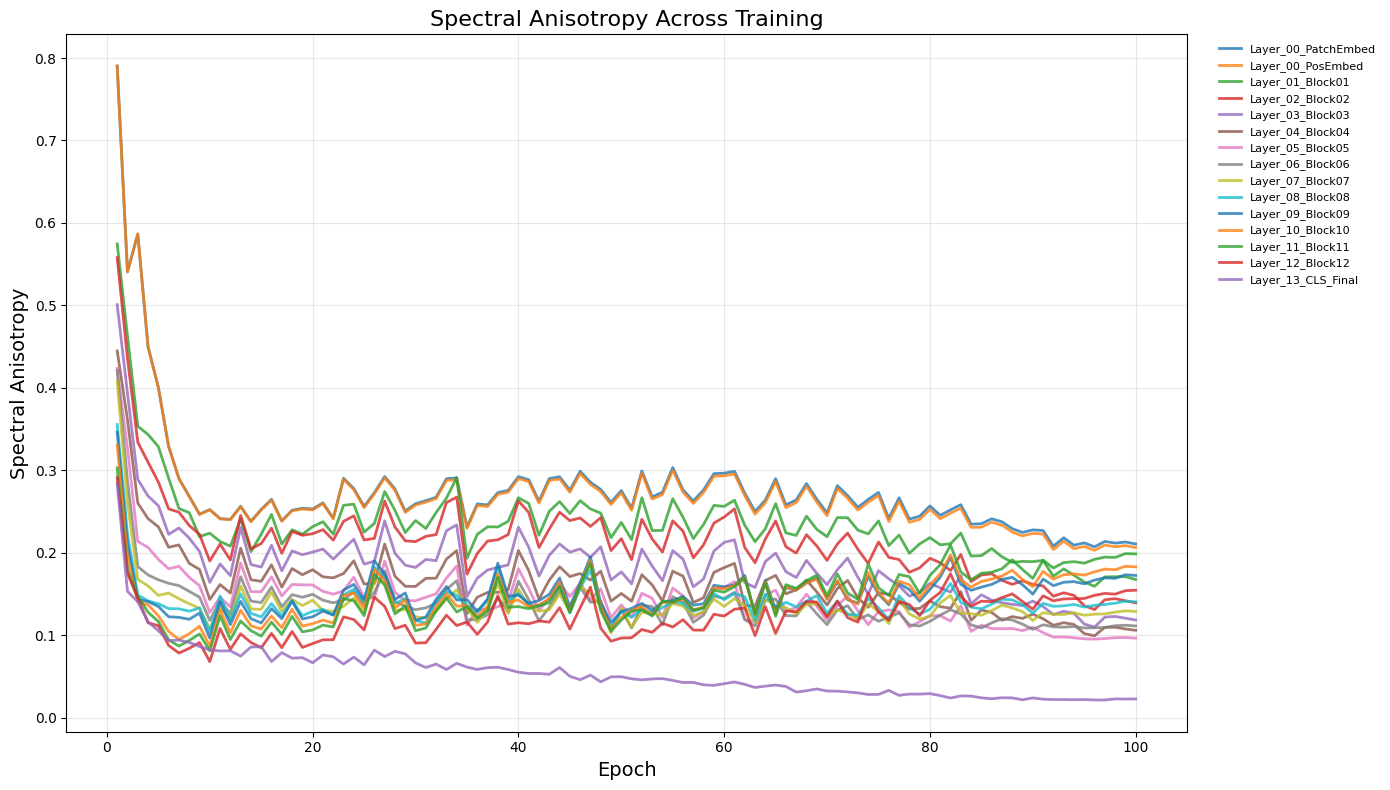}
        \caption{Spectral Anisotropy}
        \label{fig:spectral_anisotropy_layers}
    \end{subfigure}

    \vspace{0.5em}

    \begin{subfigure}[b]{0.9\textwidth}
        \centering
        \includegraphics[width=\linewidth]{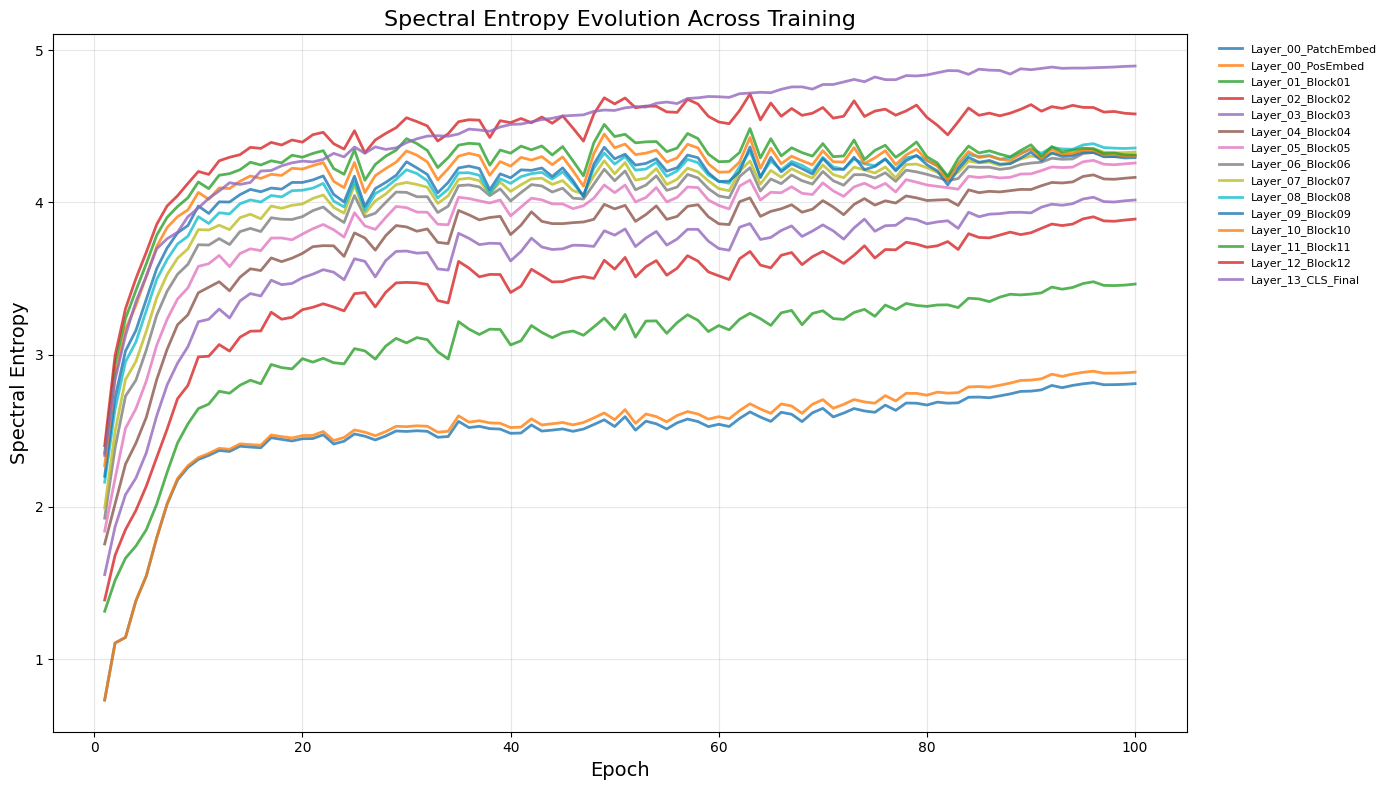}
        \caption{Spectral Entropy}
        \label{fig:spectral_entropy_layers}
    \end{subfigure}

    \caption{
    Evolution of spectral anisotropy and spectral entropy throughout training.
    Spectral anisotropy decreases across most layers, while spectral entropy increases, indicating progressively less concentrated and more distributed spectral structure.
    }
    \label{fig:anisotropy_entropy}
\end{figure*}

\subsection{Spectral Anisotropy Evolution}

\textbf{Spectral Anisotropy} was analyzed per epoch for all monitored layers. Figure~\ref{fig:spectral_anisotropy_layers} shows the evolution of Spectral Anisotropy throughout training. A consistent decrease in Spectral Anisotropy was observed across most layers of the network. Early layers exhibited the highest anisotropy values and remained comparatively more anisotropic than deeper representations throughout training. In contrast, deeper Transformer blocks showed progressively lower anisotropy values as training progressed. The strongest reduction was observed in the final CLS representation, which reached the lowest anisotropy values in the network by epoch 100. Across the model, the decrease in Spectral Anisotropy occurred alongside increases in Effective Rank, Participation Ratio, and Spectral Entropy. The observed trend indicates that variance became progressively less concentrated in a small number of dominant spectral directions during training.

\subsection{Spectral Entropy Evolution}

\textbf{Spectral Entropy} was analyzed per epoch for all monitored layers. Figure~\ref{fig:spectral_entropy_layers} shows the evolution of Spectral Entropy throughout training. A consistent increase in Spectral Entropy was observed across nearly all layers of the network. Early representations exhibited comparatively lower entropy values, while deeper Transformer layers achieved progressively higher entropy throughout training. The largest entropy values were observed in the final CLS representation. Across the network, Spectral Entropy increased substantially between initialization and epoch 100, with most layers displaying a rapid increase during the early stages of training followed by a slower rate of growth in later epoch. The observed increase in Spectral Entropy indicates that the eigenvalue distribution became progressively less concentrated throughout training. Variance was distributed across a larger fraction of the available spectral directions, resulting in a more uniform spectral profile relative to earlier training stages.
\subsection{Covariance Matrices}
\begin{figure}[h]
    \centering
    \includegraphics[width=1.0\linewidth]{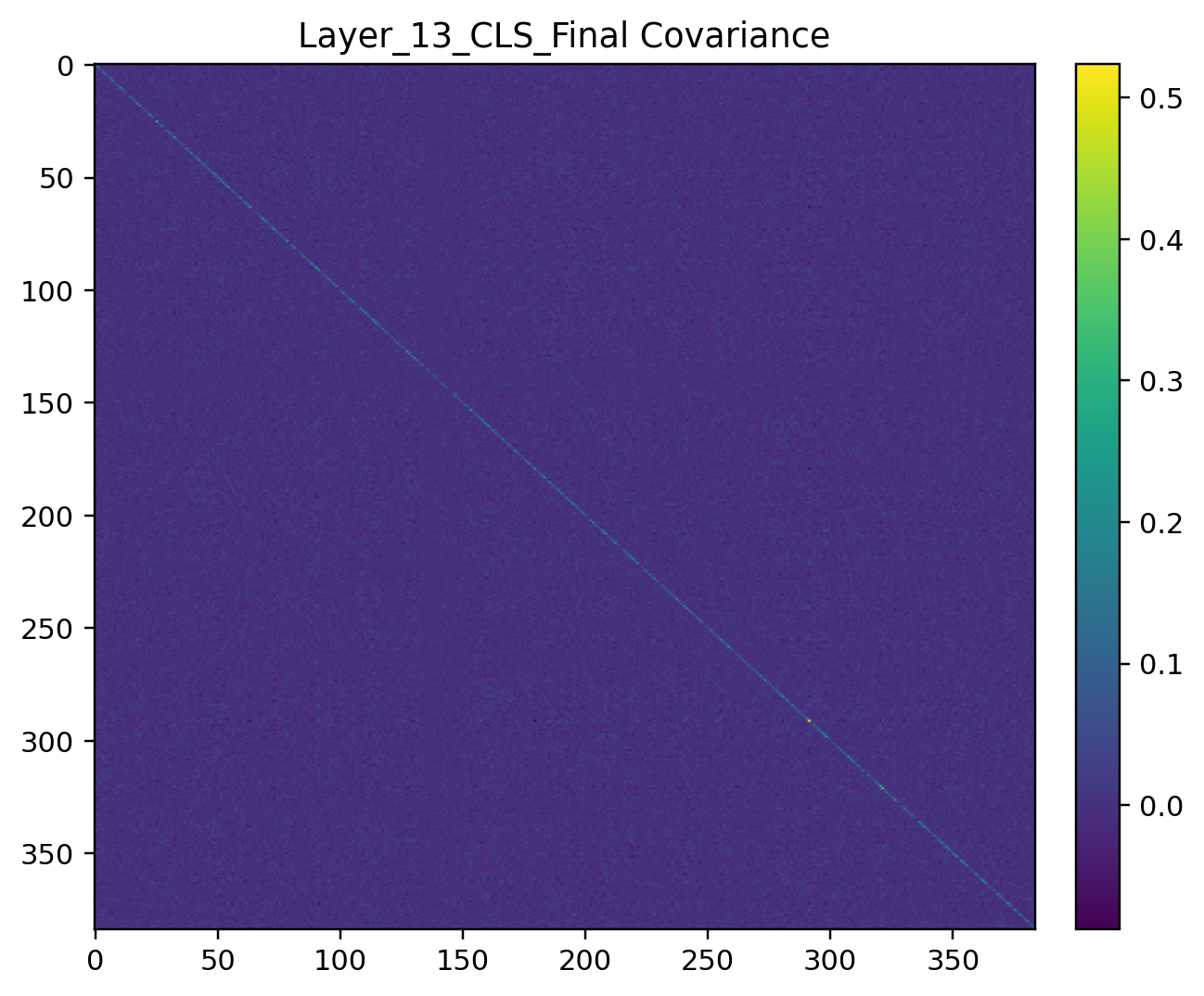}
    \caption{CLS Covariance for Epoch 100 shows how the off-diagonal correlations have become increasingly weak, pointing towards possible massive de-correlation or inherently uncorrelated feature dimensions.}
    \label{fig:covariance}
\end{figure}

Across training, the covariance structure becomes progressively more diagonal, with off-diagonal correlations becoming increasingly weak, as shown in Figure~\ref{fig:covariance}. By epoch 100, the covariance matrix exhibits a strongly diagonal structure with comparatively small off-diagonal values, similar principles were used by X-CiT~\cite{xcit} pointing to a relation between Gram Eigenvalues and covariance. The final CLS covariance matrix displays largely decorrelated feature dimensions, with variance concentrated primarily along the diagonal entries. Only a small number of localized covariance hotspots remain visible, indicating limited residual coupling between specific feature pairs. No prominent block structures or strongly correlated feature groups were observed. These observations are consistent with the increasing Effective Rank, Participation Ratio, and Spectral Entropy, as well as the decreasing Spectral Anisotropy observed throughout training.
\subsection{Spectral Evolution Across Training}

\begin{figure*}[h]
\centering
\setlength{\tabcolsep}{2pt}
\renewcommand{\arraystretch}{1.0}

\begin{tabular}{@{}ccc@{}}
\textbf{Epoch 20} & \textbf{Epoch 50} & \textbf{Epoch 100} \\[0.6em]

\includegraphics[width=0.31\textwidth]{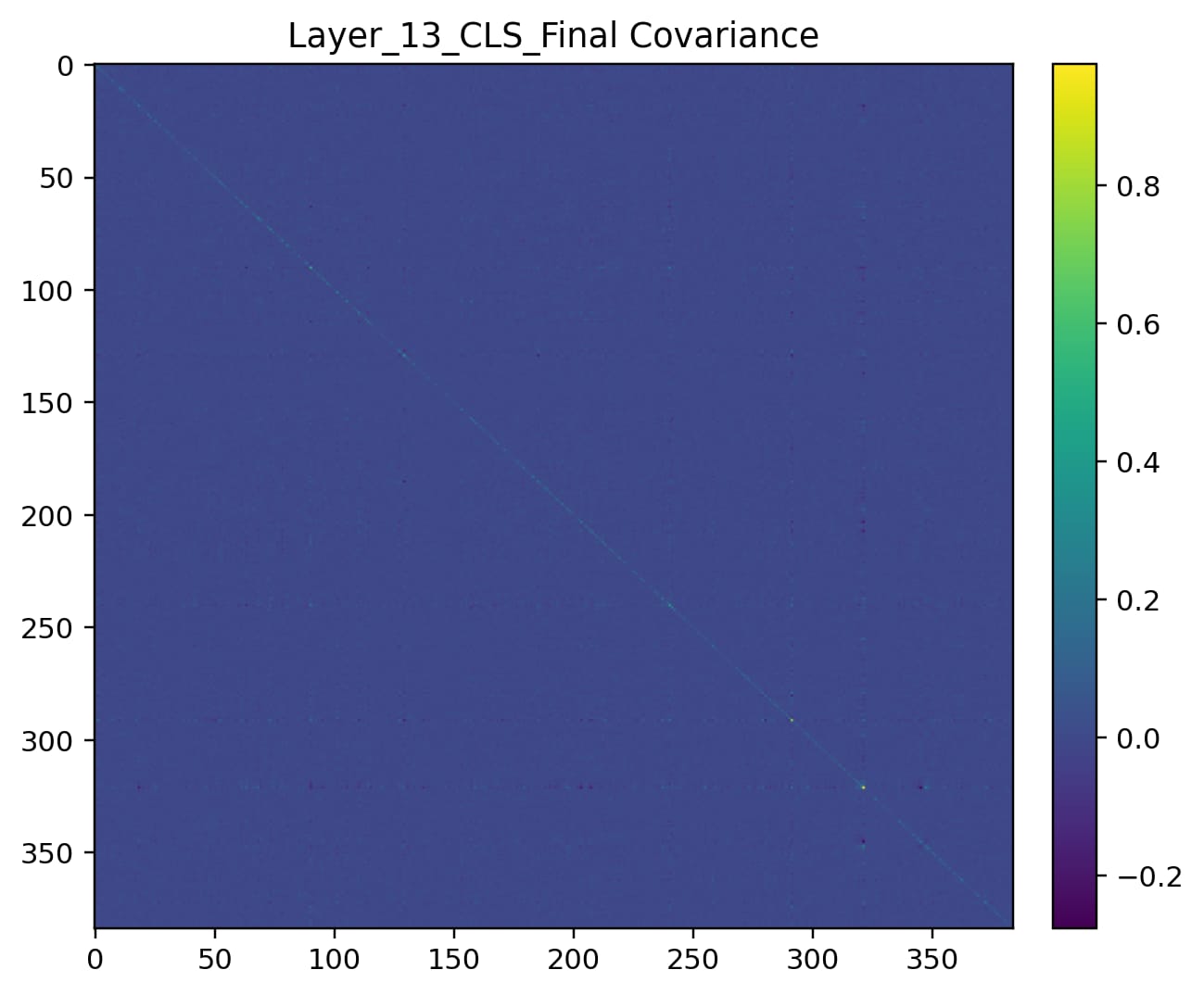} &
\includegraphics[width=0.31\textwidth]{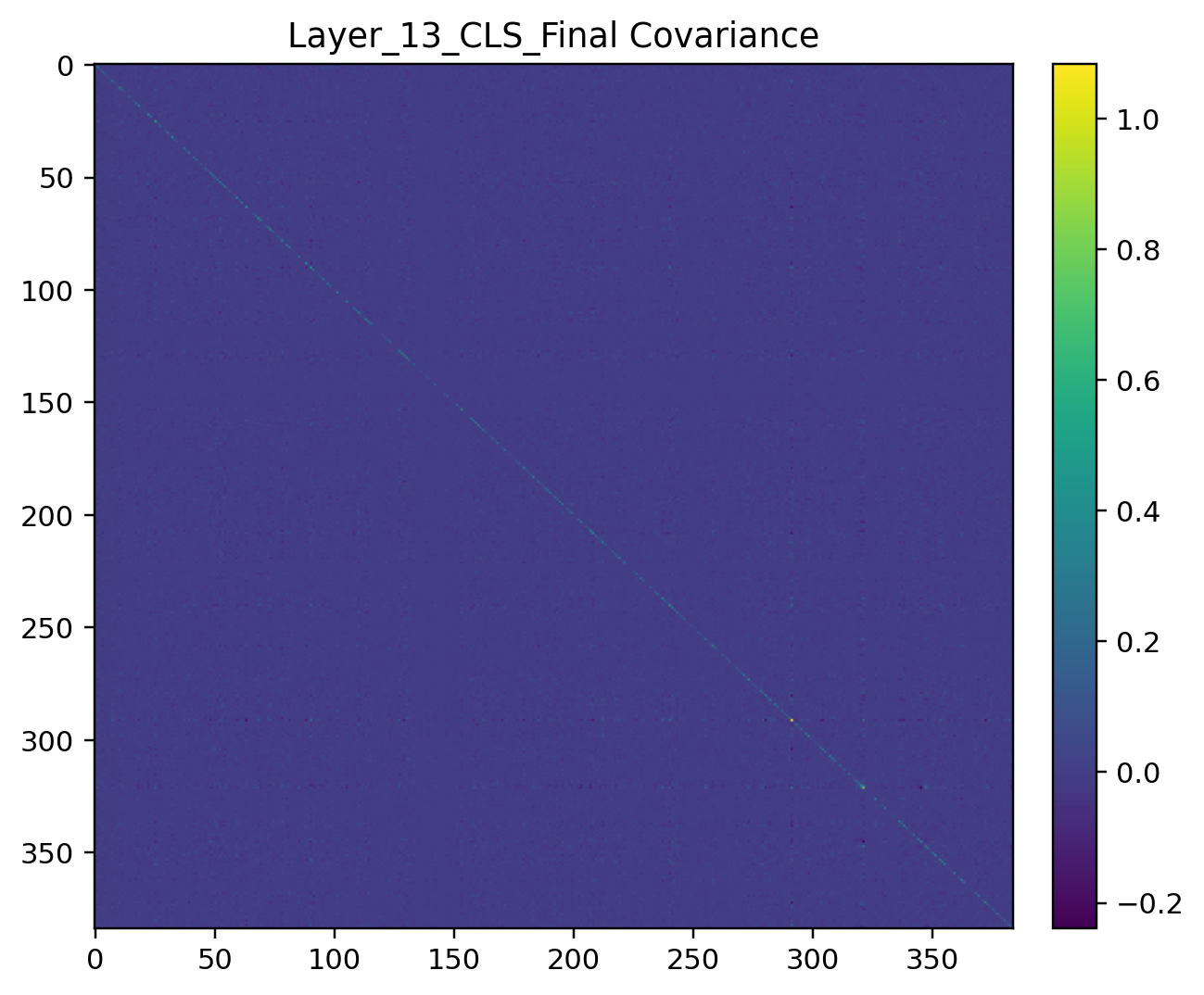} &
\includegraphics[width=0.31\textwidth]{epoch_100/covariance_heatmap.png} \\[0.6em]

\includegraphics[width=0.31\textwidth]{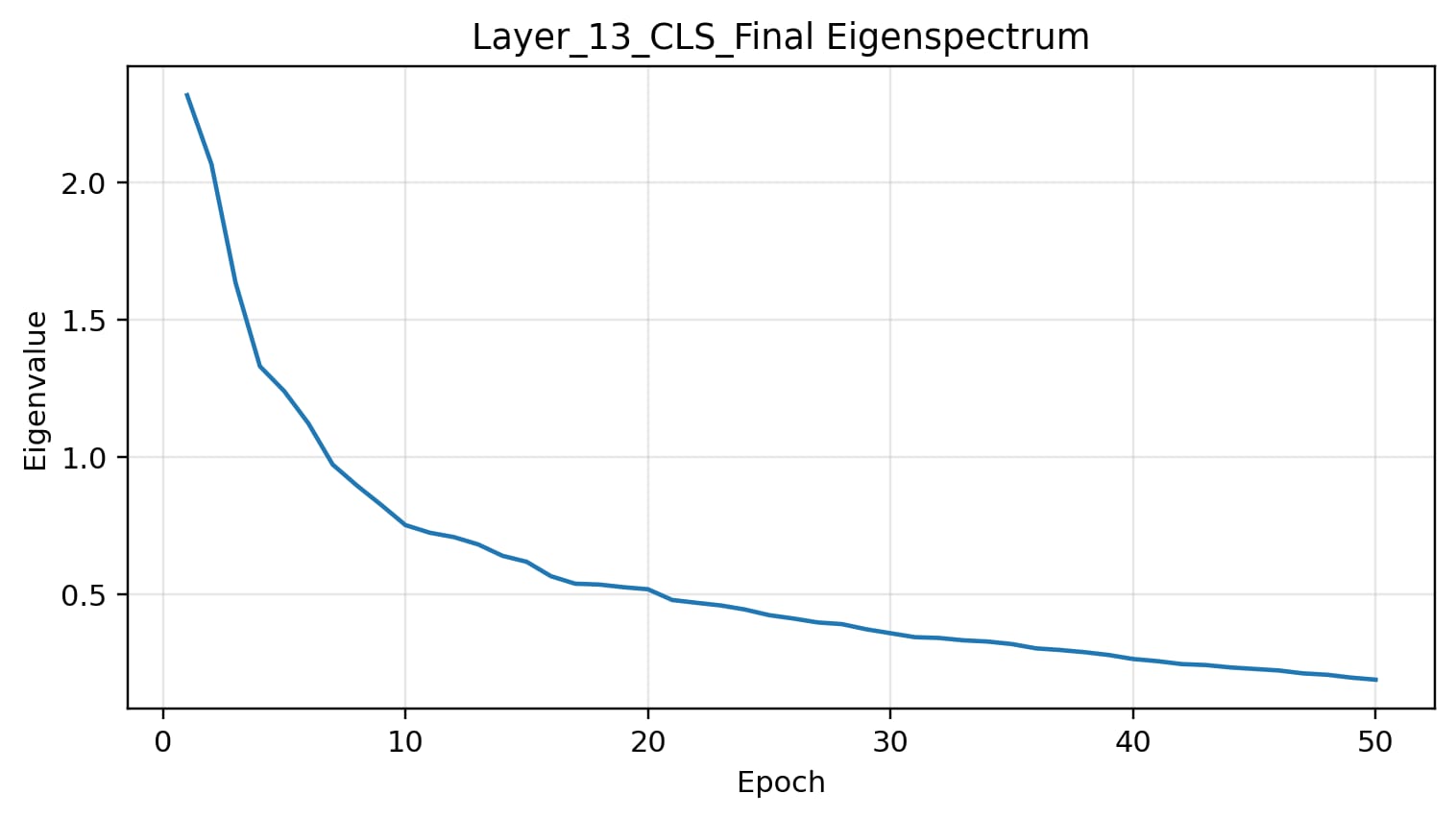} &
\includegraphics[width=0.31\textwidth]{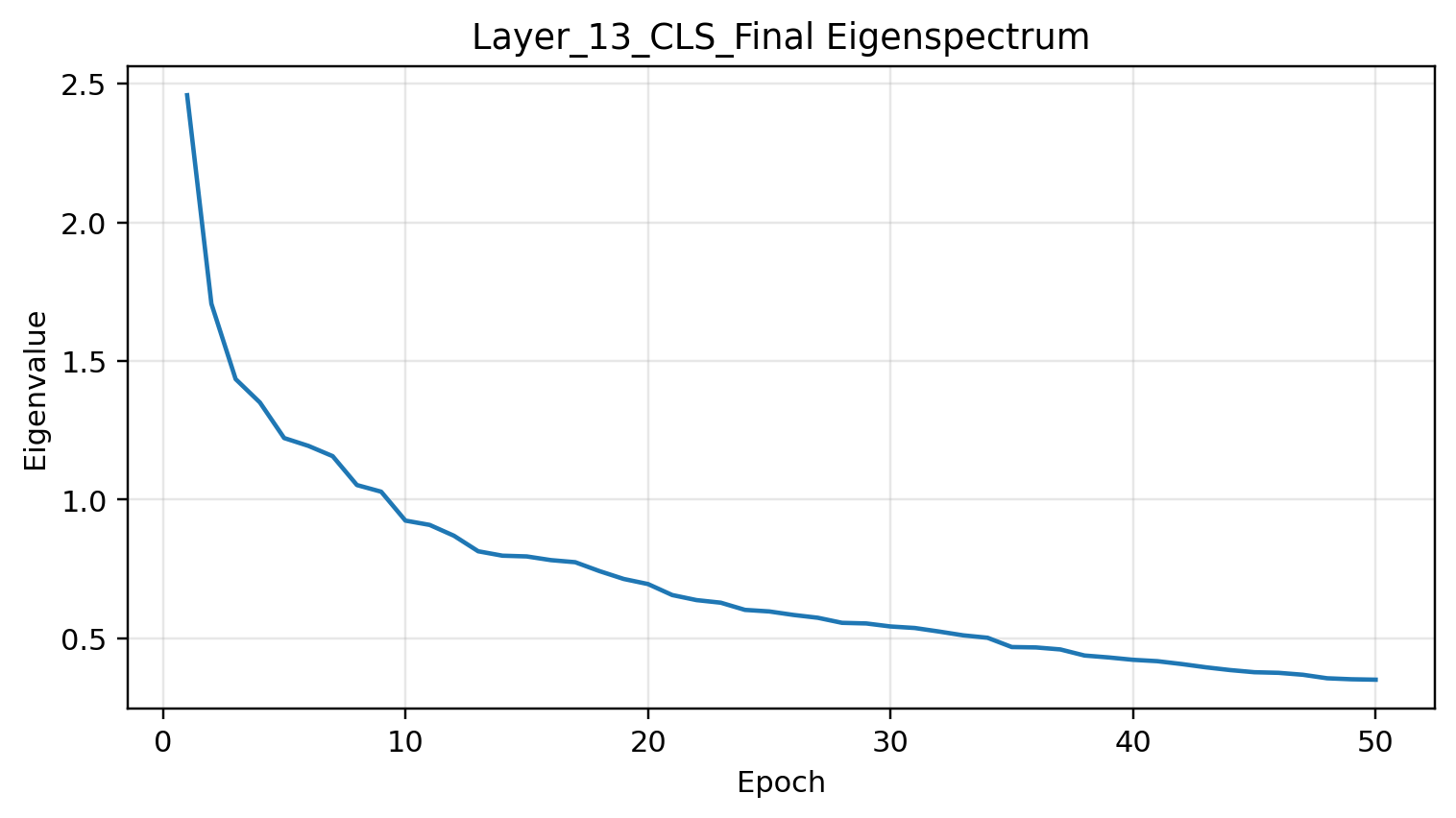} &
\includegraphics[width=0.31\textwidth]{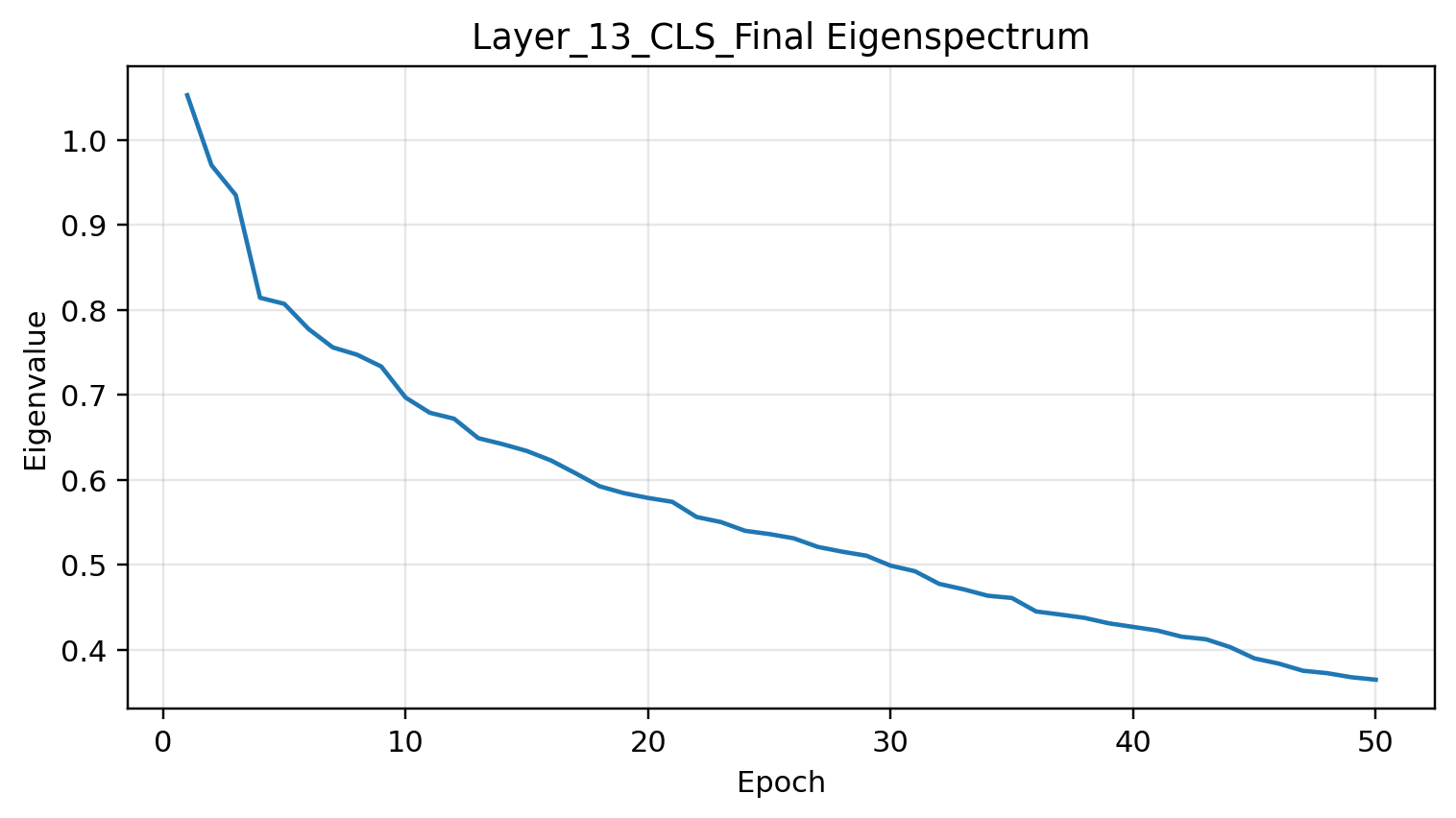} \\[0.6em]

\includegraphics[width=0.31\textwidth]{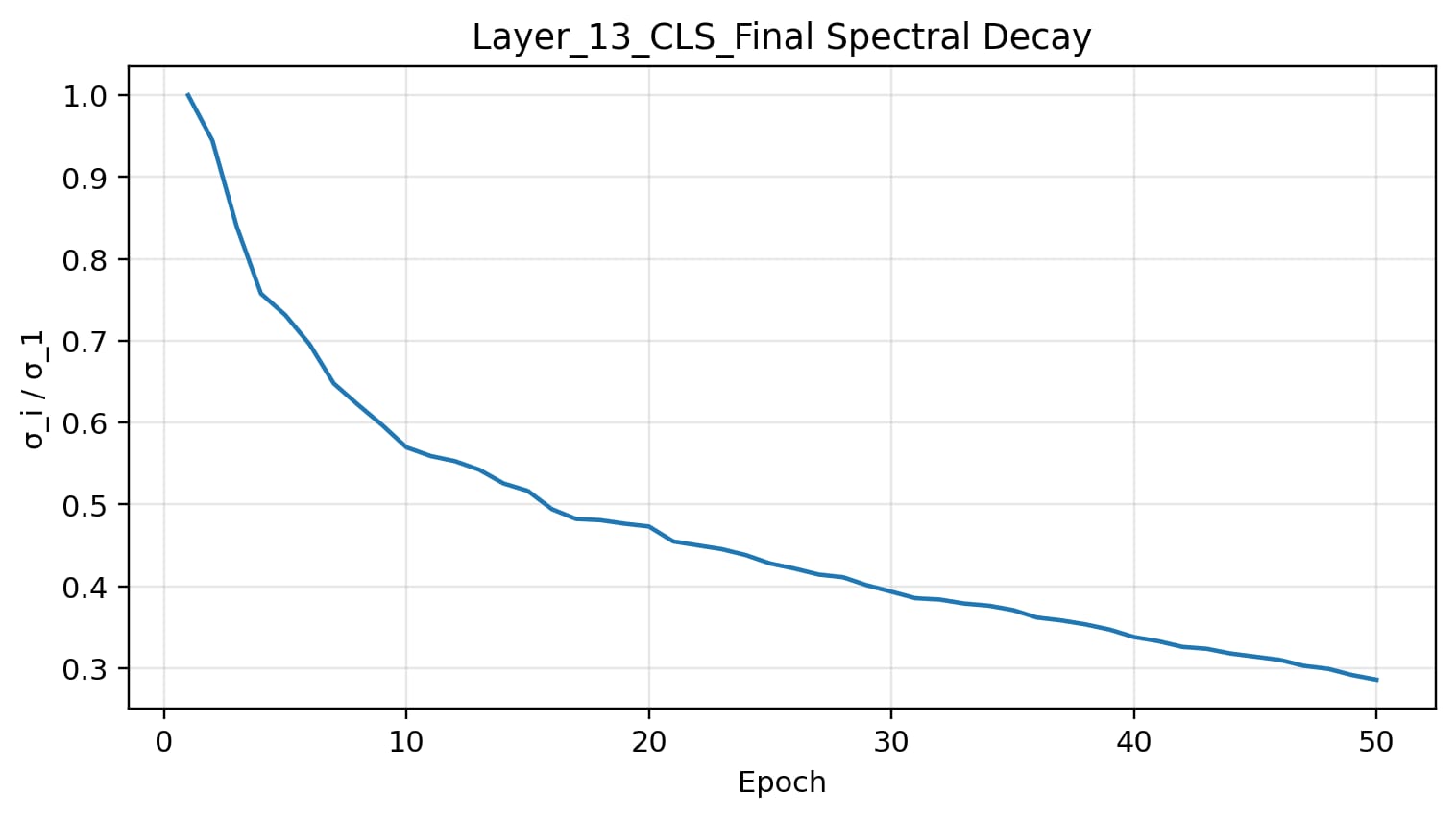} &
\includegraphics[width=0.31\textwidth]{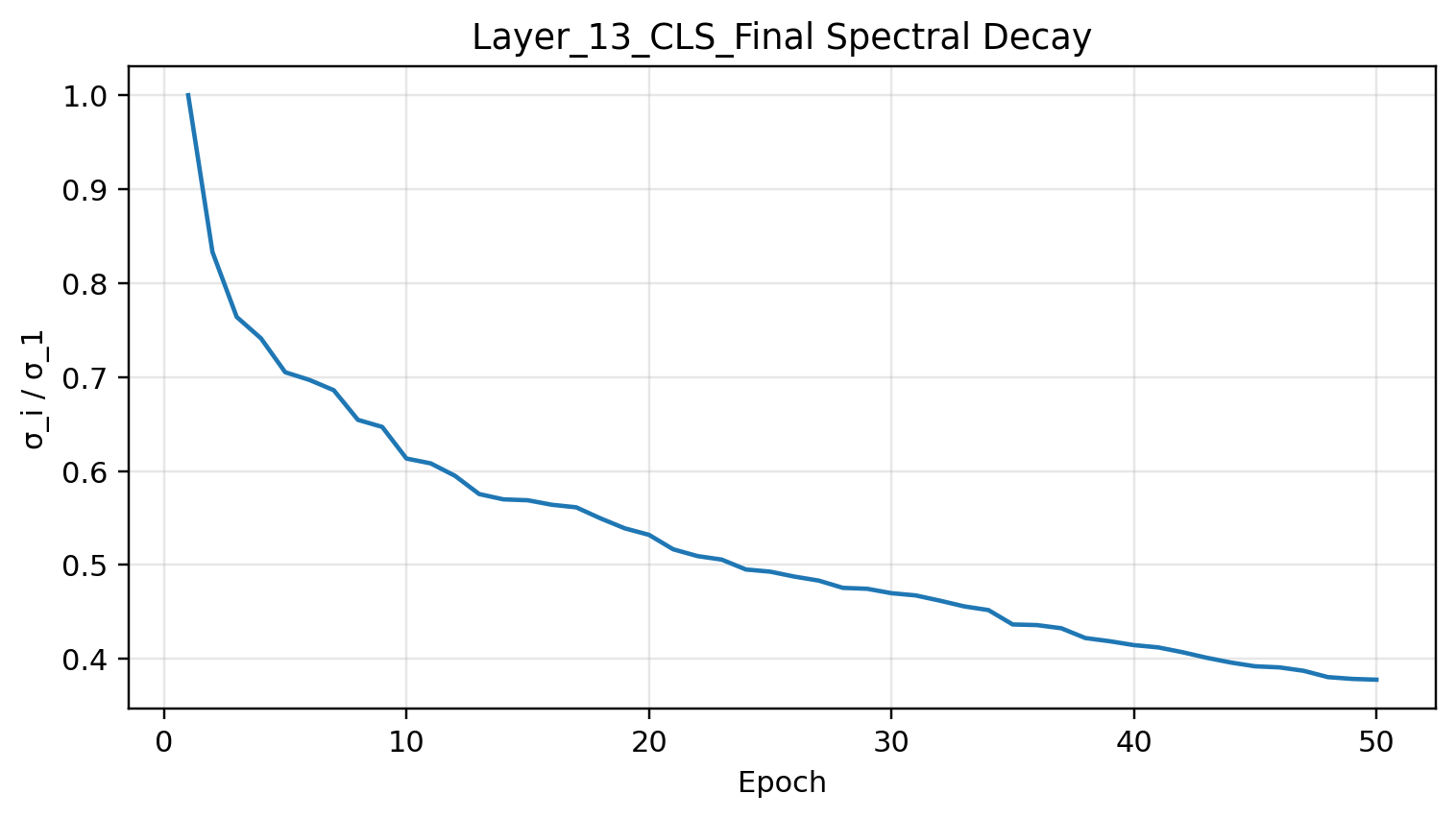} &
\includegraphics[width=0.31\textwidth]{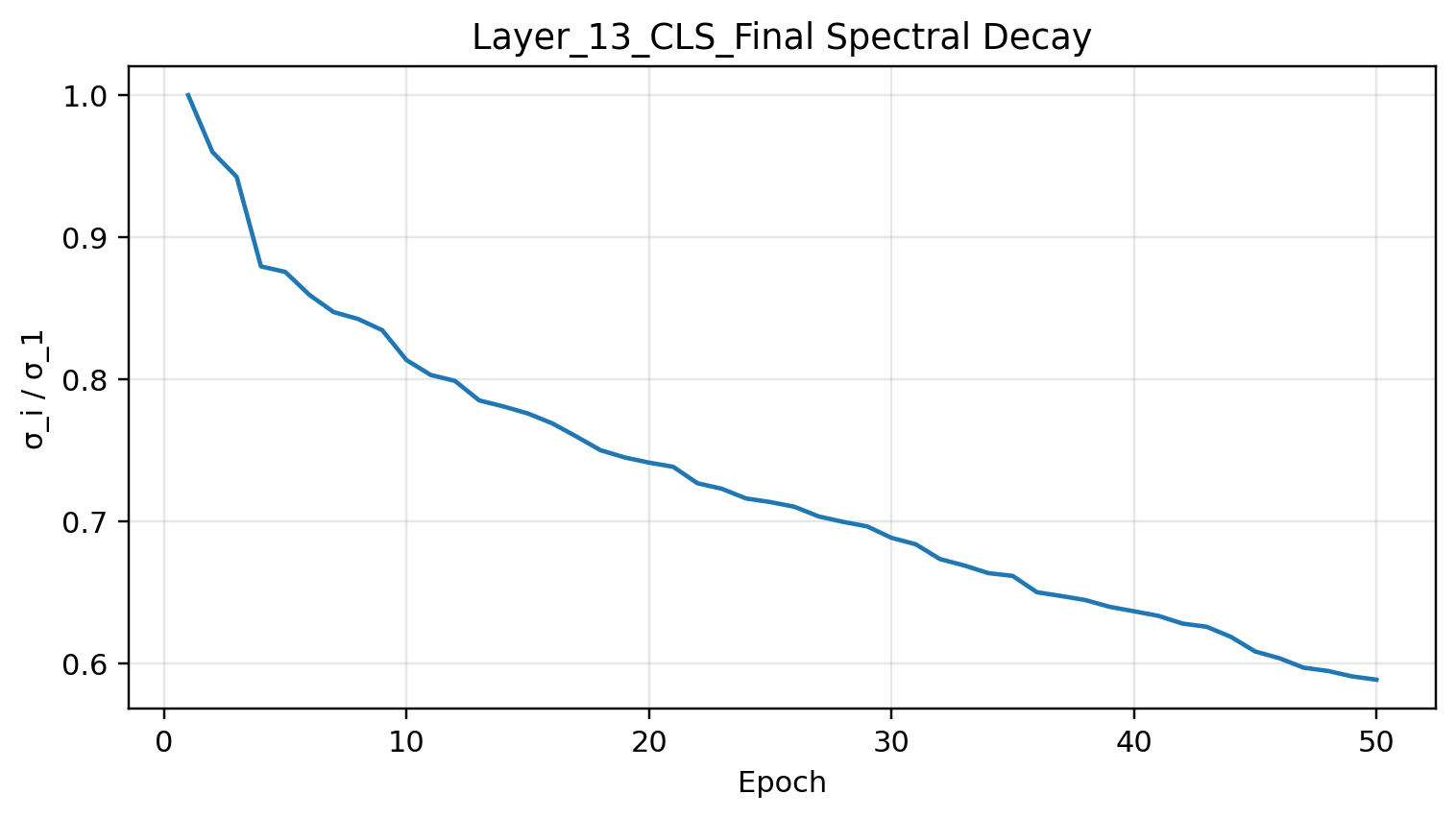} \\[0.6em]

\includegraphics[width=0.31\textwidth]{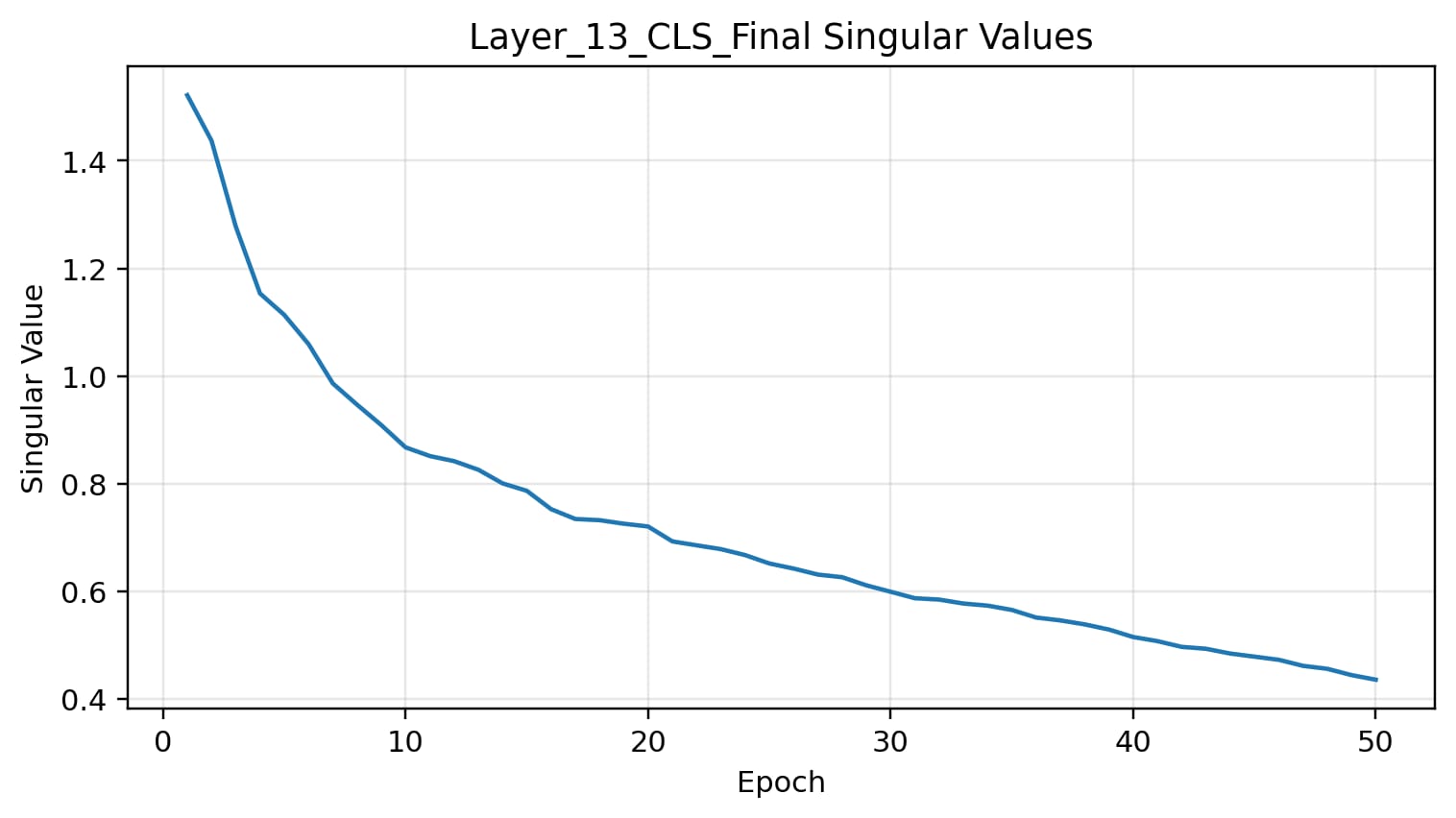} &
\includegraphics[width=0.31\textwidth]{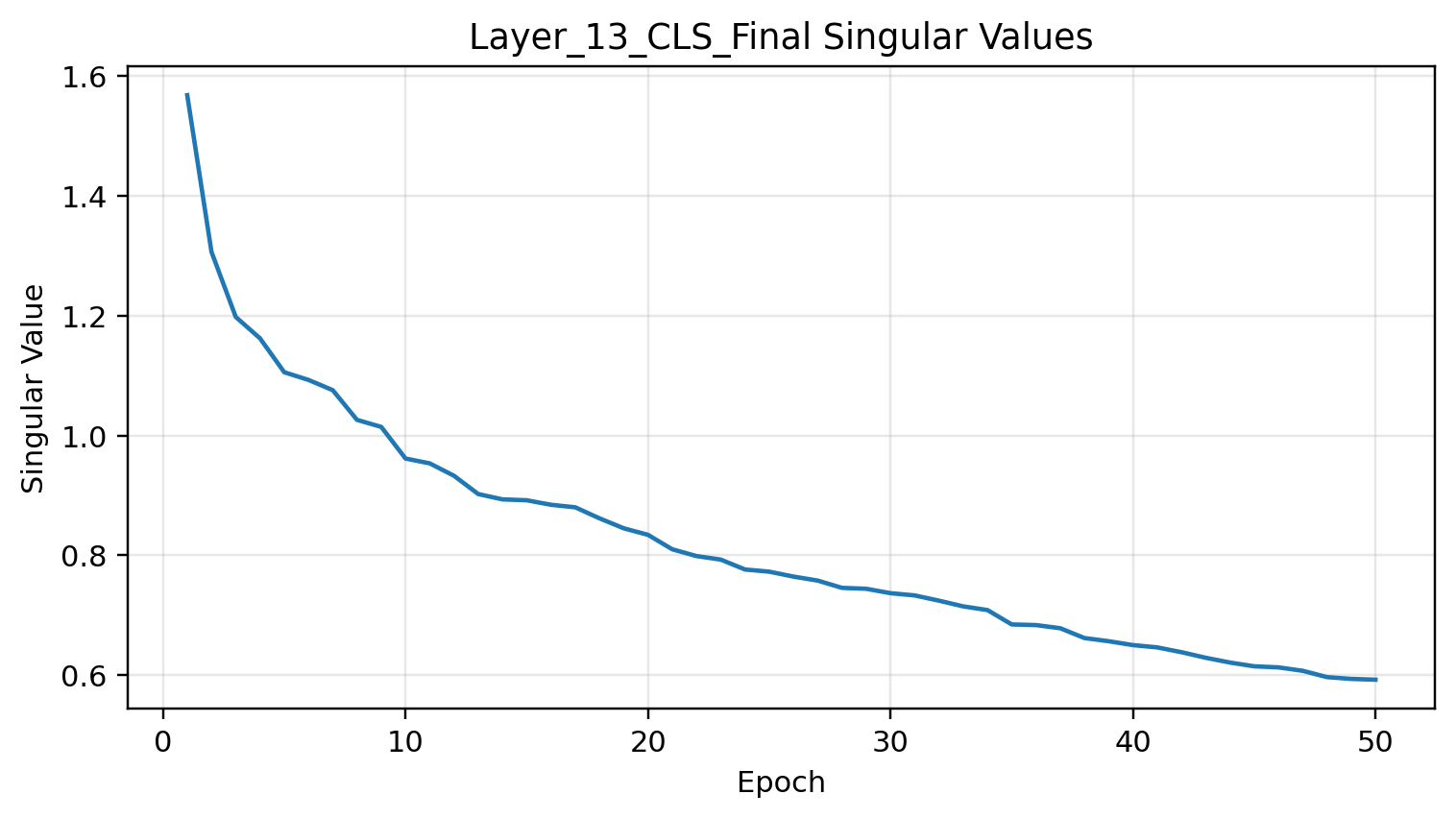} &
\includegraphics[width=0.31\textwidth]{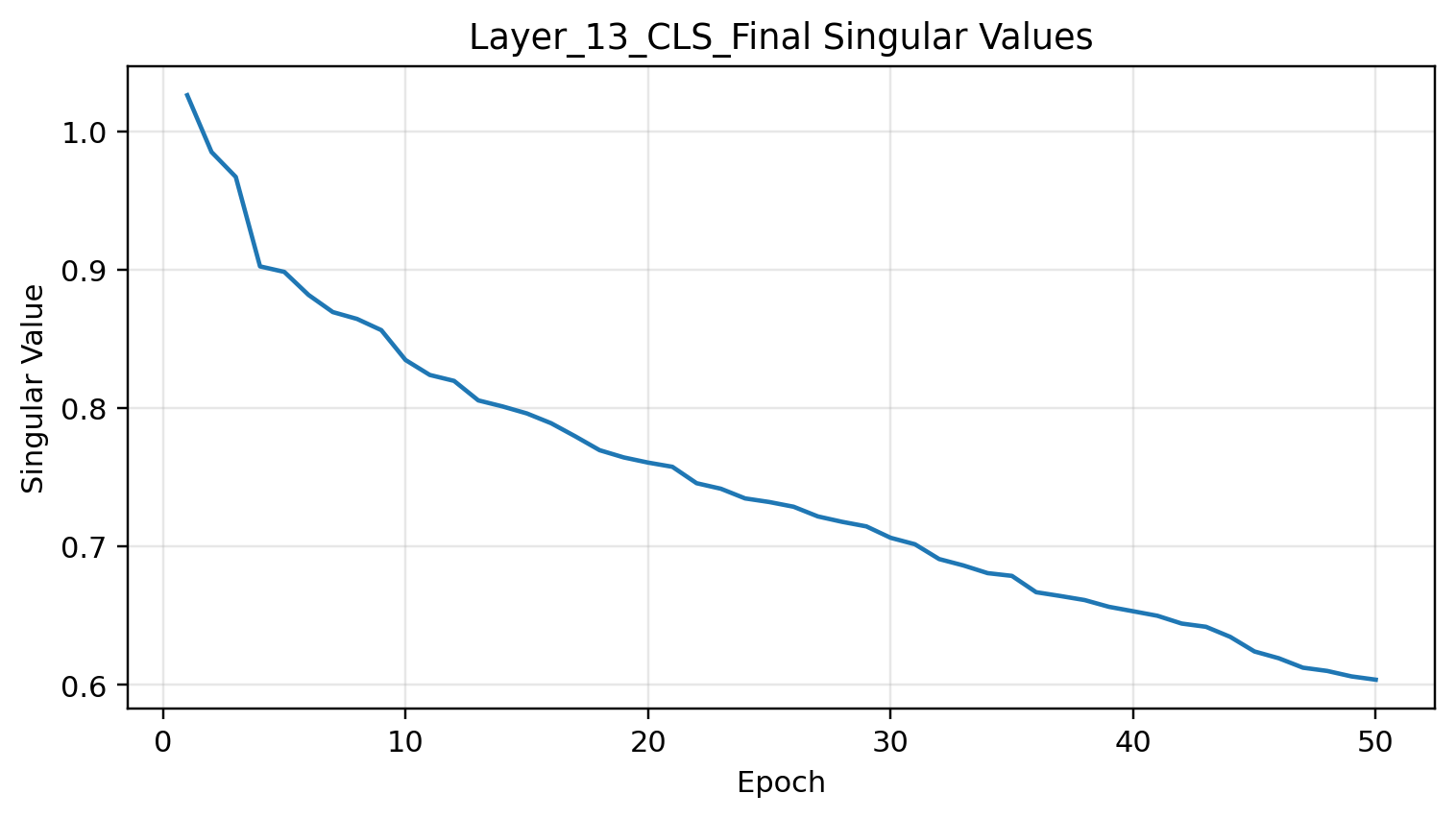}
\end{tabular}

\caption{Evolution of the final CLS representation across epochs 20, 50, and 100. Rows correspond to covariance matrices, eigenspectra, spectral decay curves, and singular value spectra, respectively. The figure shows progressive diagonalization of covariance structure, flattening of the eigenspectrum, and reduced spectral concentration over training.}
\label{fig:cls_epoch_grid}
\end{figure*}

To complement the scalar observables discussed previously, Figure~\ref{fig:cls_epoch_grid} presents the evolution of the final CLS representation at epochs 20, 50, and 100 through eigenspectra, spectral decay curves, and singular value spectra. Across training, the eigenspectrum becomes progressively flatter, with variance distributed across a larger fraction of the available eigen-directions. The dominance of the leading eigenvalues decreases relative to the remainder of the spectrum, resulting in a less concentrated spectral profile at later stages of training. A similar trend is observed in the spectral decay curves. Early training stages exhibit comparatively steeper decay, whereas later epochs display a more gradual decline. This behavior indicates that spectral energy is distributed across a broader range of representational directions as training progresses. The singular value spectra exhibit consistent behavior, becoming progressively less concentrated over time. Larger portions of the singular value distribution contribute meaningfully to the overall representation, suggesting increased utilization of the available embedding space. Collectively, these observations are consistent with the increases in Effective Rank, Stable Rank, Participation Ratio, and Spectral Entropy, as well as the decrease in Spectral Anisotropy observed throughout training. The direct spectral visualizations therefore provide qualitative support for the broader trend of progressive spectral redistribution within the final CLS representation.

\paragraph{NOTE} It is also observed from all the plotted curves that the final few layers tend to show almost identical rank, anisotropy and entropy trends

\section{Hypotheses and Discussion}

The observations presented in the previous section reveal a consistent increase in Effective Rank, Stable Rank, Participation Ratio, and Spectral Entropy, accompanied by decreasing Spectral Anisotropy and increasingly diagonal covariance structure. While TGO-I was designed as an observational framework rather than a causal study, the measured behavior motivates several hypotheses regarding the mechanisms underlying the observed geometric evolution.

\subsection{Hypothesis I: Token Diversification}

One possible explanation for the observed increase in dimensional utilization is the progressive diversification of token representations throughout training. Under this hypothesis, tokens become increasingly distinguishable within the learned feature space, causing variance to occupy a larger number of spectral directions. Such behavior would naturally lead to increases in Effective Rank, Participation Ratio, and Spectral Entropy as we saw in Figures ~\ref{fig:effective_rank_layers}~\ref{fig:spectral_entropy_layers}. The analyses performed in TGO-I do not directly measure token similarity or token-level redundancy. Consequently, this hypothesis remains unverified within the current observatory. Future TGO observatories incorporating token cosine similarity, token covariance analysis, and token trajectory measurements may help TO evaluate this possibility.

\subsection{Hypothesis II: Semantic Expansion}

As observed in Figure~\ref{fig:spectral_anisotropy_layers}, there is a expansion in the number of directions that the model explores which leads us to our second possibility, that is training - progressively discovers and encodes a larger number of semantically meaningful feature directions. Under this interpretation, the observed increase in dimensional utilization reflects the emergence of increasingly rich representational structure rather than simple redistribution of variance. The substantial growth observed in the final CLS representation may be consistent with this interpretation, as the CLS token functions as a global aggregation mechanism for information throughout the network. However, TGO-I does not directly measure semantic content or feature meaning. Additional studies involving class separation, representation similarity, and semantic probing would be required to investigate this hypothesis.

\subsection{Hypothesis III: Redundancy Reduction}

A third possibility is that the observed spectral evolution reflects a progressive reduction in representational redundancy. Unlike the previous hypotheses, which attribute the increase in dimensional utilization to either token diversification or the emergence of new semantic directions, this interpretation suggests that training reorganizes existing information into a more efficient representational basis. As shown by the increasingly diagonal covariance structure, feature dimensions become progressively less correlated throughout training, indicating reduced statistical dependence between learned features. Under this hypothesis, the network is not necessarily encoding substantially more information, but rather distributing information across dimensions in a less redundant manner. Such behavior would naturally produce representations in which variance is spread more evenly across the available feature space and fewer dimensions carry duplicated information. While the covariance visualizations are qualitatively consistent with this interpretation, TGO-I does not directly measure redundancy or information overlap between features. Future observatories incorporating feature correlation analysis, mutual information estimation, and redundancy metrics will be required to determine whether redundancy reduction is a primary driver of the observed geometric evolution.

\subsection{Summary}

At present, TGO-I provides evidence for progressive spectral redistribution within Vision Transformer representations but does not establish the underlying causal mechanism. Token diversification, semantic expansion, and redundancy removal remain plausible explanations for the observed behavior. Distinguishing between these hypotheses constitutes a primary objective of future Transformer Geometry Observatory investigations.
\section{Conclusion}

TGO-I introduced the first observatory within the Transformer Geometry Observatory (TGO) framework and investigated the spectral evolution of Vision Transformer representations throughout training. By analyzing covariance structure, Effective Rank, Stable Rank, Participation Ratio, Spectral Entropy, Spectral Flatness, Spectral Anisotropy, eigenspectra, and singular value spectra across all layers and training epochs, a consistent pattern of progressive spectral redistribution was observed. Across training, Effective Rank, Stable Rank, Participation Ratio, and Spectral Entropy increased, while Spectral Anisotropy decreased. Covariance matrices became increasingly diagonal, indicating progressively weaker feature correlations. These observations suggest that representation geometry evolves toward a more distributed and less spectrally concentrated structure as training progresses. The strongest effects were observed within the final CLS representation, which consistently exhibited the highest dimensional utilization and lowest anisotropy within the network. In contrast, Patch Embedding and Positional Embedding layers remained comparatively stable throughout training, suggesting that the majority of geometric evolution occurs within Transformer processing blocks and the final global representation.

While TGO-I establishes several robust empirical observations, the underlying mechanism remains unresolved. Token diversification, semantic expansion, and redundancy removal remain plausible explanations for the observed behavior. Distinguishing between these competing hypotheses requires additional observatories capable of directly measuring token dynamics, representational similarity, information flow, and optimization geometry. The primary contribution of TGO-I is therefore not a definitive explanation of transformer behavior, but the establishment of a systematic observational framework capable of generating reproducible measurements, identifying unexpected phenomena, and motivating future mechanistic investigations of transformer learning dynamics.

\section{Future Work}

TGO-I represents only the first component of a broader research program aimed at systematically characterizing transformer learning dynamics. Future observatories will investigate complementary aspects of transformer behavior and seek to transform observational findings into mechanistic explanations.

\subsection{TGO-II: Representation Similarity Observatory}

Several spectral observables exhibited clustering behavior in which groups of adjacent Transformer layers converged toward nearly identical metric trajectories. While spectral similarity alone does not imply representational equivalence, this observation motivates direct representation similarity analysis using methods such as CKA, SVCCA, and PWCCA. These experiments will determine whether spectrally similar layers perform redundant computations or occupy distinct representational subspaces.

\subsection{TGO-III: Token Dynamics Observatory}

The token diversification hypothesis proposed in TGO-I cannot be evaluated using covariance spectra alone. Future work will therefore investigate token-level dynamics through cosine similarity analysis, token covariance structure, token trajectory visualization, token drift measurements, and local manifold analysis. These experiments will determine whether the observed spectral expansion is accompanied by increasing token diversity.

\subsection{TGO-IV: Attention Geometry Observatory}

Attention mechanisms remain one of the least directly observed components of transformer learning dynamics. Future observatories will analyze attention entropy, head specialization, attention sparsity, routing structure, token dominance, and information flow across layers. These measurements may reveal whether attention contributes to representation expansion, compression, or redundancy reduction.

\subsection{TGO-V: Optimization Geometry Observatory}

Future work will investigate gradient geometry, Hessian geometry, curvature evolution, and loss landscape dynamics throughout training. These measurements may reveal phase transitions, stability regions, and optimization bottlenecks that are not visible through representation geometry alone.

\subsection{Long-Term Objective}

The long-term objective of TGO is to establish a unified framework for observing, measuring, visualizing, and explaining transformer learning dynamics. By integrating representation geometry, optimization geometry, information geometry, attention geometry, positional geometry, and dynamical systems analysis, future observatories aim to identify hidden bottlenecks, unexpected behaviors, and mechanistic explanations that may inform the design of future transformer architectures, sparse models, and neuromorphic systems.

\bibliographystyle{IEEEtran}
\bibliography{references}

\end{document}